\documentclass[sigconf,final]{acmart}

\usepackage{multirow,multicol}

\AtBeginDocument{%
  \providecommand\BibTeX{{%
    \normalfont B\kern-0.5em{\scshape i\kern-0.25em b}\kern-0.8em\TeX}}}

\setcopyright{acmcopyright}
\copyrightyear{2021}
\acmYear{2021}
\acmDOI{10.1145/1122445.1122456}

\acmConference[ICMI '21]{Montreal '21: Proceedings of the 23rd ACM International Conference on Multimodal Interaction}
\acmBooktitle{ }
\acmPrice{15.00}
\acmISBN{978-1-4503-XXXX-X/18/06}

\DeclareMathOperator\BiGRU{BiGRU}
\DeclareMathOperator\avgpool{avgpool}

\usepackage{soul}
\begin{document}

\title{Bi-Bimodal Modality Fusion for Correlation-Controlled Multimodal Sentiment Analysis}

\author{Wei Han}
\affiliation{%
  \institution{ISTD, SUTD}
  \city{Singapore}
  \country{Singapore}
}
\email{wei_han@mymail.sutd.edu.sg}

\author{Hui Chen}
\affiliation{%
 \institution{ISTD, SUTD}
  \city{Singapore}
  \country{Singapore}
}
\email{hui_chen@mymail.sutd.edu.sg}

\author{Alexander Gelbukh}
\affiliation{%
  \institution{CIC, IPN}
  \city{Mexico City}
  \state{CDMX}
  \country{Mexico}
}
\email{gelbukh@gelbukh.com}

\author{Amir Zadeh}
\affiliation{%
 \institution{LTI, CMU}
  \city{Pittsburgh}
  \state{PA}
  \country{USA}
}
\email{abagherz@cs.cmu.edu}

\author{Louis-philippe Morency}
\affiliation{%
  \institution{LTI, CMU}
  \city{Pittsburgh}
  \state{PA}
  \country{USA}
}
\email{morency@cs.cmu.edu}

\author{Soujanya Poria}
\affiliation{%
 \institution{ISTD, SUTD}
  \city{Singapore}
  \country{Singapore}
}
\email{sporia@sutd.edu.sg}

\newcommand{\modelname}{BBFN}
\newcommand{\modelfull}{Bi-Bimodal Fusion Network}

\newcommand{\vm}{visual}
\newcommand{\am}{acoustic}
\newcommand{\tm}{text}
\newcommand{\Vm}{Visual}
\newcommand{\Am}{Acoustic}
\newcommand{\Tm}{Text}

\newcommand{\N}{\mathcal{N}}
\newcommand{\hw}[1]{\textcolor{green}{[Henry: #1]}} 

\begin{abstract}
Multimodal sentiment analysis aims to extract and integrate semantic information collected from multiple modalities to recognize the expressed emotions and sentiment in multimodal data.
This research area's major concern lies in developing an extraordinary fusion scheme that can extract and integrate key information from various modalities.
However, previous work is restricted by the lack of leveraging dynamics of independence and correlation between modalities to reach top performance.
To mitigate this, we propose the \textit{Bi-Bimodal Fusion Network} (BBFN), a novel end-to-end network that performs fusion (relevance increment) and separation (difference increment) on pairwise modality representations. The two parts are trained simultaneously such that the combat between them is simulated. The model takes two bimodal pairs as input due to the known information imbalance among modalities. 
In addition, we leverage a gated control mechanism in the Transformer architecture to further improve the final output.
Experimental results on three datasets (CMU-MOSI, CMU-MOSEI, and UR-FUNNY) verifies that our model significantly
outperforms the SOTA. The implementation of this work is available at \url{https://github.com/declare-lab/multimodal-deep-learning}.
\end{abstract}  


\maketitle
\section{Introduction}
\begin{figure}[t]
    \centering
    \includegraphics [
        trim=0.8cm 0cm 0.2cm 0cm, clip,
        width=0.48\textwidth]{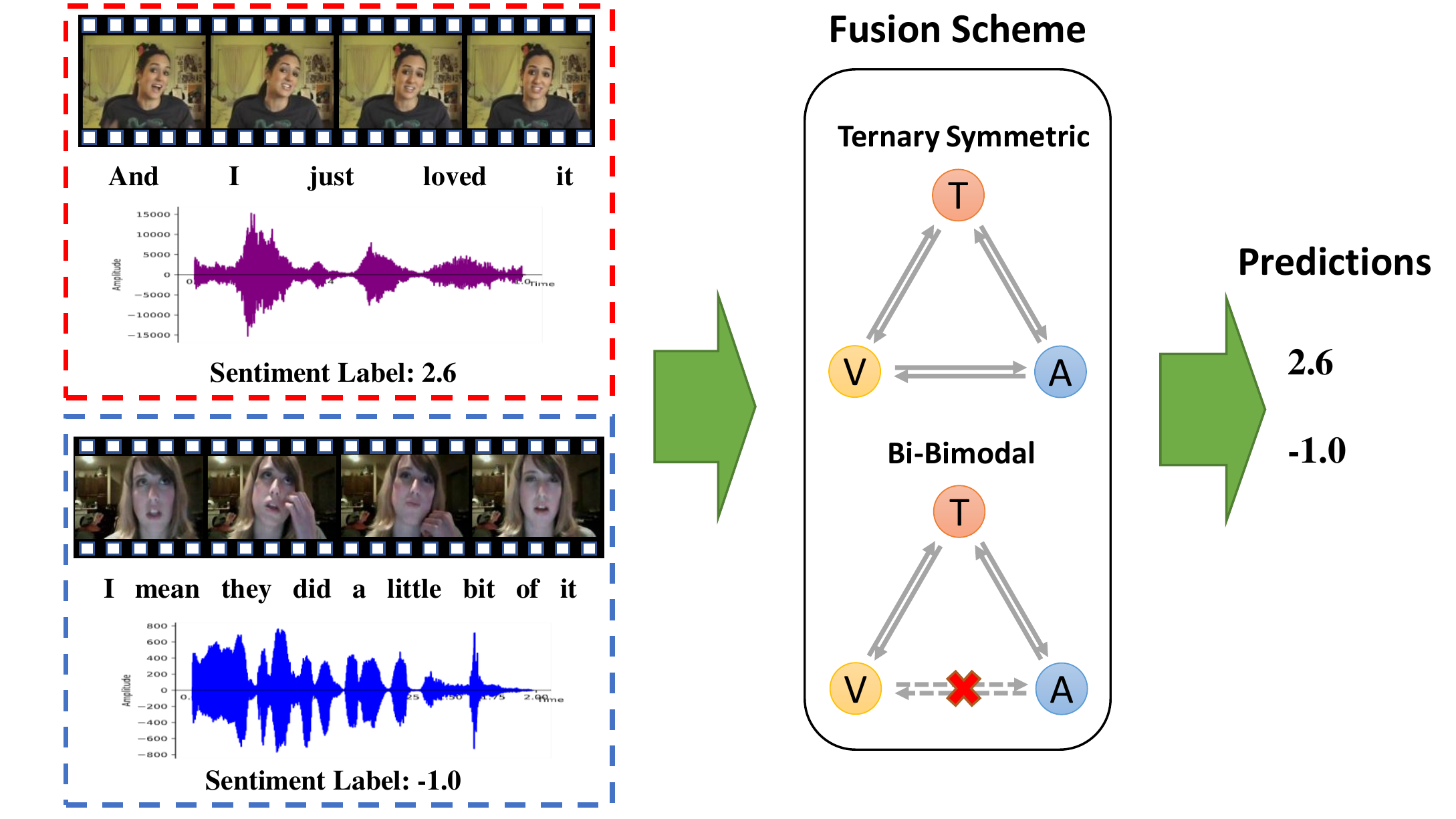}
    \caption{Task formulation of the multimodal sentiment analysis and two types of fusion schemes. The displayed data are sampled from CMU-MOSI dataset.}
    \label{Problem description}
\end{figure}
With the unprecedented prevalence of social media in recent years and the availability of phones with high-quality cameras, we witness an explosive boost of multimodal data, such as video clips posted on different social media platforms. 
Such multimodal data consist of three channels:
\vm\ (image), \am\ (voice), and transcribed linguistic (text) data.
Different modalities in the same data segment are often complementary to each other, providing extra cues for semantic and emotional disambiguation~\cite{ngiam2011multimodal}. 
For example, the phrase ``apple tree" can indicate what the blurred red fruit on the tree is in an image, and a smiling face can clarify that some seemingly impolite words are a friendly joke.
On the other hand, the three modalities usually possess unique statistical properties that make them to some extent mutually independent---say, one modality can stay practically constant while the other one exhibits large changes~\citep{srivastava2014multimodal}. 
Accordingly, a crucial issue in multimodal language processing is how to integrate heterogeneous data efficiently.
A good fusion scheme should 
extract and integrate meaningful information from multiple modalities while preserving their mutual independence.

In this paper, we focus on the problem of multimodal sentiment analysis (MSA). 
As Fig. \ref{Problem description} suggests, given data from multiple modality sources, the goal of MSA is to exploit fusion techniques to combine these modalities to make predictions for the labels. 
In the context of emotion recognition and sentiment analysis, multimodal fusion is essential since emotional cues are often spread across different modalities~\citep{atrey2010multimodal}. 
Previous research in this field~\citep{zadeh2017tensor,tsai2019multimodal,hazarika2020misa} mostly adopted ternary-symmetric architectures, where bidirectional relationships in every modality pair are 
modeled in some way. 
However, as it has been pointed out by several past research~\citep{chen2017multimodal, tsai2019multimodal, sun2020learning},
the task-related information is not evenly distributed between the modalities.
The architectures that do not account for this difference in the relative importance of the three modalities fail to fuse them correctly, which 
degrades the model's performance.

To address this issue, we introduce a fusion scheme that we call \textit{\modelfull} (\modelname) to balance the contribution of different modality pairs properly.
This fusion scheme, consisting of two bi-modal fusion modules, is quite different from traditional ternary symmetric one; see Fig. \ref{Problem description}.
Since it has been empirically shown that the text modality is most significant \citep{tsai2019multimodal, pham2019improving},
our model takes two text-related modality pairs, TV (\tm-\vm) and TA (\tm-\am),
as respective inputs for its two bimodal learning modules.
Then it iteratively impels modalities to complement their 
information through interactive learning with their corresponding counterparts.
To ensure fairness in the bidirectional learning process for both modalities, the two learning networks in each model should be identical. 
The basic framework of our model is layers of stacked Transformers, which have been proven efficient in multimodal learning \citep{yu2020improving}. 

However, a new problem arises in our implementation.
As fusion proceeds, the representation vectors of the fusion results involved with a modality pair tend to become closer in the hidden space; we call it \textit{feature space collapse}.
Moreover, the repeated structures of transformers in the stacked architecture exacerbate this trend, impairing the mutual independence between different modalities present in the multimodal data--a crucial property for the feasibility of multimodal fusion. 
To tackle this problem, we introduce in our \modelname\ the layer-wise feature space separator, as a local regularizer that divides 
the feature space of different modalities in order to assure mutual independence between modalities. 

We evaluated our model on two subtasks of MSA---sentiment intensity prediction and humor recognition---using three datasets: CMU-MOSI, CMU-MOSEI, and UR-FUNNY.
Our experimental results show that our model outperforms state-of-the-art models on almost all metrics. 
Moreover, ablation study and further analysis show the effectiveness of our architecture.

Our contributions 
can be summarized as follows:
    \begin{itemize}
        \item \textbf{Bi-bimodal fusion:} We introduce a novel fusion scheme for MSA, 
        which
        consists of two Transformer-based bimodal learning modules, each one taking a modality sequence pair as input and performing a progressive fusion locally in its two modality complementation modules.
        \item \textbf{Regularization:} To enforce modality representations to be 
        unique and different from each other, 
        we use a modality-specific feature separator, which implicitly clusters homogeneous representations and splits heterogeneous ones apart in order to maintain mutual independence between modalities.
        \item \textbf{Control:} 
        We 
        introduce a gated control mechanism to enhance the Transformer-based fusion process.
    \end{itemize}

\section{Related Work}
In this section, we briefly overview related work in 
MSA and
multimodal fusion.

\subsection{Multimodal Sentiment Analysis}
Multimodal sentiment analysis (MSA) mainly focuses on integrating multiple resources, such as acoustic, visual, and textual information, to comprehend varied human emotions~\citep{morency2011towards}. In the past few years, deep neural networks have been employed in learning multimodal representation in sentiment analysis, such as Long Short-Term Memory (LSTM), which is used to model long-range dependencies from low-level multimodal features~\citep{wollmer2013youtube,chen2017multimodal} 
and SAL-CNN~\citep{wang2017select}, which utilizes a select-additive learning procedure to improve the generalizability of trained neural networks. 

Most of the previous work in this area focuses on early or late fusion. For example, \citet{zadeh2017tensor} proposed a Tensor Fusion Network, which blends different modality representations at a deeper layer.
As attention mechanism becomes more and more popular, \citet{zadeh2018multi} modified LSTM with a novel Multi-attention Block to capture interactions between modalities through time. 
Also, \citet{gu2018multimodal} introduced a hierarchical attention strategy with word-level fusion to classify utterance-level sentiment. 
Moreover, \citet{akhtar2019multi} presented a deep multi-task learning framework to jointly learn sentiment polarities and emotional intensity in a multimodal background. \citet{rahman2020integrating}~directly worked on BERT and designed functional gates to control the dataflow of one modality from other two modalities.

\citet{pham2019found} proposed a method that cyclically translates between modalities to learn robust joint representations for sentiment analysis. 
More recently, \citet{hazarika2020misa} attempted to factorize modality features in joint spaces to effectively capture commonalities across different modalities and reduce their gaps. 
\citet{tsai2020multimodal} proposed a Capsule Network-based method to dynamically adjust weights between modalities. 
Most of these works incorporate interactions in every modality pairs into their design.
In contrast, our model only includes two pairs involving a common central modality.

\subsection{Multimodal Language Learning}
\paragraph{Correlation-based Approach}
Correlation has been learned as an important metric for objects showing concurrently. There are many previous works that include this item for various purposes.~\citet{sun2019multi,sun2020learning} directly optimized over a correlation-related DCCA loss to learn multimodal representations useful for downstream tasks.~\citet{mittal2020m3er} instead used correlation as a selection criteria to guide multimodal data to orderly form a union representation. Although all these works took correlation into account, they ignored the importance of modality's independence and the competition between the two opponents.

\paragraph{Alignment-based Approach}
Alignment is the process to map signals of different sampling rates to the same frequency. Early multimodal alignment approaches~\citep{zadeh2016multimodal,blikstein2016multimodal} usually firstly chose a target frequency and then calculated the frames in each modality that need mapping to that position. Some classical loss functions like CTC~\citep{graves2006connectionist} and their variants are widely used to facilitate alignment improvement. Thanks to the advent attention mechanism, the Transformer architecture shows state-of-the-art performance in multiple disciplines in both text and visual fields~\citep{vaswani2017attention,dosovitskiy2020image}. Unlike traditional alignment routines, attention naturally formulates a point-to-point mapping between two modalities, which is called ``soft alignment'' and has been proven effective in more general cases of multimodal representation learning and feature fusion. For example, \citet{yu2020improving} designed a Unified Multimodal Transformer to jointly model text and visual representations in the NER problem. Moreover, \citet{tsai2019multimodal} employed the Transformer to model as well as align sequences from visual, textual, and acoustic sources. 
Our fusion architecture is built on Transformer, but performs fusion in a progressive manner with feature space regularization and fine-grained gate control. 

\paragraph{Application in Other Tasks}
Besides multimodal sentiment analysis, multimodal learning has been applied in many other language tasks, such as Machine Translation (MT)~\citep{specia2016shared,elliott2018adversarial, yao2020multimodal,huang2020unsupervised}, Named Entity Recognition (NER)~\citep{moon2018multimodal,zhang2018adaptive,lu2018visual,yu2020improving}, and parsing~\citep{shi2018learning,shi2019visually,zhao2020visually}. 

\newcommand{\modala}{main}
\newcommand{\modalb}{comp}
\newcommand{\modalafull}{main}
\newcommand{\modalbfull}{complementary}
\newcommand{\Modalafull}{Main}
\newcommand{\Modalbfull}{Complementary}

\section{Methodology}
In this section, we first briefly define the problem and then describe our \modelname\ model.
\paragraph{Task Definition}
The task of MSA aims to predict sentiment intensity, polarity, or emotion label of given multimodal input (video clip). 
The video consists of three modalities: $t$ (\tm), $v$ (\vm) and $a$ (\am), which are 2D tensors 
denoted by $M_t\in \mathbb{R}^{T_t\times d_t}$, $M_v\in \mathbb{R}^{T_v\times d_v}$, and $M_a \in \mathbb{R}^{M_a\times d_a}$, where $T_{m}$ and $d_{m}$ represent sequence length (e.g., number of frames) and feature vector size of modality $m$. 

\subsection{Overall Description}

\begin{figure*}
    \centering
    \includegraphics[trim=0 0.1cm 0 0, width=0.8\textwidth]{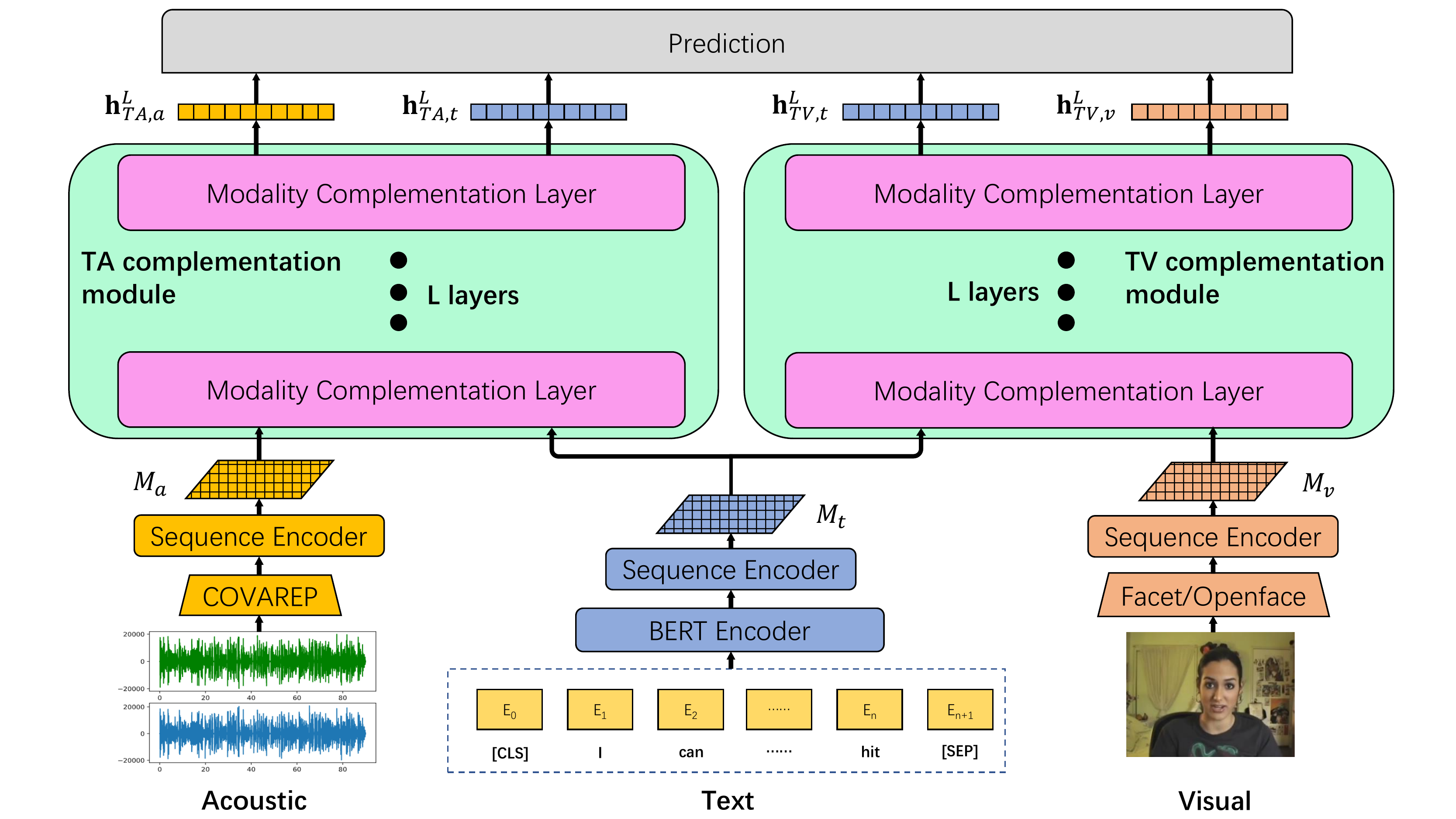}
    \caption{Overview of our \modelfull\ (\modelname). It 
    learns two text-related pairs of representations, TA and TV, by causing each pair of modalities to complement mutually. Finally, the four (two pairs) head representations are concatenated to generate the final prediction.}
    \label{Fig.GTR}
\end{figure*}

The overview of our model is shown in Fig. \ref{Fig.GTR}.
It consists of two modality complementation modules, each accomplishing a bimodal fusion process in its two fusion pipelines.
After receiving context-aware representations from the underlying modality sequence encoders, bimodal fusion proceeds iteratively through stacked complementation layers. 

The feature space separator is another key idea of our model.
Each modality has its own feature representations. However, in a deep neural network, when these unique unimodal representations propagate through multiple layers, their mutual independence can be compromised, i.e., they may not be as separable as they were 
initially;
we call this \textit{feature space collapse}. 
The separability of the unimodal representations and their mutual independence is necessary for multimodal fusion; otherwise, one modality can hardly learn something new from its counterparts through heterogeneous attention on respective hidden representations.
Accordingly, we enforce these representations to preserve more modality-specific features to prevent them from collapsing into a pair of vectors with similar distributions.

Finally, the conventional heterogeneous Transformer purely uses a residual connection to combine attention results and input representations without any controlled decision made for the acceptance and rejection along the hidden dimension of these vectors. 
Instead,
we incorporate a gated control mechanism in the multi-head attention of the Transformer network, which also couples the feature separator and transformer fusion pipelines. 

\subsection{Modality Sequence Encoder}
\label{BERT}
We encode all modality sequences to guarantee a better fusion outcome in the subsequent modality complementation modules.
\paragraph{Word Embedding} We use the Transformer-based pre-trained model, BERT ~\citep{devlin2019bert} as the text encoder. The raw sentence $S=(w_1,\cdots,w_n)$ composed of word indices is firstly concatenated with two special tokens---{\fontfamily{qcr}\selectfont [CLS]} at the head and {\fontfamily{qcr}\selectfont [SEP]} at the tail and then fed into the encoder to generate contextualized word embeddings, as the input of text modality $M_t=(m_0,m_1,...,m_{n+1})$.

\paragraph{Sequence Encoder}
The input modality sequences $M_{m}$, $m\in\{t,v,a\}$, are essentially time series and exhibit temporal dependency. 
We use a single-layer bidirectional gated recurrent unit (BiGRU)~\citep{chung2014empirical} followed by a linear projection layer to capture their internal dependency and cast all the hidden vectors to the same length for the convenience of further processing. 
The resulting sequences are
\begin{align}
X^0_{m}=(x_{m,0}^0,x_{m,1}^0,...,x_{m,n+1}^0), 
\label{eq:X}
\end{align}
where $m \in \{t,v,a\}$ denotes 
a modality. These outputs serve as the initial inputs to the modality complementation module. 

\subsection{Modality Complementation Module}
\label{Transformer Layer}
In the modality complementation module, the modality representation pairs exchange information with their counterparts to ``complement" the missing cues when passing through the multimodal complementation layers that interconnect two fusion pipelines with layer-wise modality-specific feature separators.
We further improve the attention-based fusion procedure by adding a gated control mechanism to enhance its performance and robustness.
The module is built in a stacked manner to realize an iterative fusion routine.

\paragraph{Modality-Specific Feature Separator} 
To maintain mutual independence among these modalities, we leverage the regularization effect exerted 
by a discriminator loss, which tells how well the hidden representations can be distinguished from their counterparts in the same complementation module. 
A straightforward solution for a separator according to prior work~\citep{hazarika2020misa,sun2020learning} is to add some geometric measures to the total loss as regularization term, such as (1)~euclidean distances or cosine correlation or (2)~distribution similarity measures such as KL-Divergence or Wasserstein distance along the hidden dimension. 
However, we chose the discriminator loss because---unlike
geometric measures, which directly use hidden vectors---it is a parametric method, so it is more suitable to be coalesced into the entire model. 

Namely,
after collecting the outputs from the previous complementation layer $X_m^{i-1}=(x_{m,0}^{i-1}, x_{m,1}^{i-1},...,x_{m,n+1}^{i-1})$, we encode the sequence with a bidirectonal GRU and then apply an average pooling to 
acquire sequence-level hidden representations:
\begin{align*}
    \bar{\mathbf{h}}_m^i &= \avgpool(\mathbf{h}_m^i) = \avgpool(\BiGRU(X_m^{i-1};\theta_m^i)).
\end{align*}
where $\theta_m^i$ are the parameters of the $\BiGRU$ in the $i$th layer. 
Here we choose BiGRU as the intermediate sequence encoder because with fewer parameters, in our experiments it provided results comparable with those of BiLSTM.
Note that until now we just described the data flow of a single modality. In a complementation module, at each layer $i$ there are always two pipelines that generate the sequences of hidden representations concurrently for two different modalities, $m_1$ and $m_2$. 

Next we want to separate the possibly entangled intermediate modality representations.
Different from previous works that rely on explicit distance maximization, we train a classifier to discern which modality these representations come from. A straightforward approach is directly fed all of them into the classifier, but it may occur serious issue: random noises in sequence representations cause the classifier to pay worthless effort on trivial features. We introduce a simple group strategy to mitigate this issue, which applies average operation on representations in the same group to generate a smoother representation. Specifically, after setting group size as $K$, the representation for the $r^{th}$ ($r=1,2,...,N/K$) group is:
\begin{equation}
    \tilde{\textbf{h}}_{m}^{i,r} = \frac{1}{K}\sum_{j=1+(r-1)\times K}^K \bar{\textbf{h}}_m^{i,j}
\end{equation}
\begin{equation}
    \hat{c}^i_r, \hat{c}^i_{r+N_b/K}= D_i(\tilde{\mathbf{h}}_{m_1}^{i,r}, \tilde{\mathbf{h}}_{m_2}^{i,r}).
\end{equation}
We leave a short explanation about how this reduces noise here. Suppose vectors possessing similar property (i.e. from the same modality in context) can be fitted with a set of gaussian distribution $\{\N(\mu_1,\sigma_1),\N(\mu_2,\sigma_2),\N(\mu_3,\sigma_3),...,\N(\mu_n,\sigma_n)\}$ and corresponding weights $\{w_1,w_2,w_3,...,w_n\}$. According to the rule for the summation of gaussian distributions, we have
\begin{equation*}
    e\sim~\N\left(\sum_n w_n\mu_n, \sqrt{\sum_n w_n^2  \sigma_n^2}\right)=\N(\mu,\sigma)
\end{equation*}
By introducing the grouping trick, for each group representation the new expectation term holds constant while the variance term turns to
\begin{equation*}
    \sigma_g = \frac{\sqrt{\sum_n w_n^2  \sigma_n^2}}{K} = \sigma/K 
\end{equation*}
which decreases as group size increases.

The discriminators are distinct in every layer because of the diverse manifestations of the same modality in the semantic space as fusion progresses which thus requires different sets of parameters to discern. 
We calculate the Binary Cross Entropy between predictions and their corresponding pseudo labels that are automatically generated during training time as the discriminator loss:
\begin{equation*}
    \mathcal{L}^i_{sep} = -\frac{K}{2N_b}\sum_{r=1}^{2N_b/K}\left(c^i_r\log{\hat{c}}^i_r+(1-c^i_r)\log(1-\hat{c}^{i}_r)\right),
\end{equation*}
where $N_b$ is the batch size and $j$ represents the $j$-th sample.

\paragraph{Gated Complementation Transformer (GCT)} 
The main body of the modality complementation module is the Gated Complementation Transformer, which are stacked into two pipelines to form the symmetric structure.
For convenience of explanation, 
we will call 
the modality that keeps forwarding in the same fusion pipeline inside a complementation module the \textit{\modalafull\ modality}, denoted by $\modala$, and 
the modality that joins bimodal fusion in one pipeline but comes as an external source from the other pipeline, the \textit{\modalbfull\ modality}, denoted by $\modalb$.
Noted that distinguishing \modalafull\ and \modalbfull\ modalities makes sense only in the context of a specified pipeline.

The cross-modal fusion process occurs mainly at the multi-head attention operation, which we found to show suboptimal performance due to the lack of information flow control.
To improve it in a fine-grained and controllable way, we introduce two gates: the retain gate $\mathbf{g}_r$, which decides how much proportion of the target modality's components to be kept forwarding, and the compound gate $\mathbf{g}_c$, which decides how much proportion of compounded components to be injected to the target modality. 

We generate these two gate signals from the sequential representation of the two modalities in the same layer:
\begin{align*}
    \mathbf{g}_r^i = \sigma(\mathbf{W}^{i,\modala}_r[\bar{\mathbf{h}}_{\modala}^i\Vert\bar{\mathbf{h}}_{\modalb}^i]), \\
    \mathbf{g}_c^i = \sigma(\mathbf{W}^{i,\modala}_c[\bar{\mathbf{h}}_{\modala}^i\Vert\bar{\mathbf{h}}_{\modalb}^i]),
\end{align*}
where $\mathbf{W}_*\in \mathbb{R}^{2d\times d}$ is the projection matrix and~$\Vert$ represents concatenation. 
After 
receiving
the query $Q^i=\mathbf{W}_Q^iX^i_{\modala}$, 
key $K^i=\mathbf{W}_K^iX^i_{\modalb}$ and 
value $V^i=\mathbf{W}_V^iX^i_{\modalb}$, these gates are then employed on the multi-head attention to limit the information flow of the residual block as a part of bimodal combination:
\begin{align*}
    \textbf{m}^i &= \text{MH-ATT}(Q^i, K^i, V^i), \\
    \tilde{X}^{i}_{\modala} &= \text{LN}(\mathbf{g}_c^i \odot \mathbf{m}^i + \mathbf{g}_r^i \odot X^i_{\modala}),
\end{align*}
where MH-ATT represents multi-head attention, $\odot$ means component-wise multiplication  and LN is layer normalization. 
Next, the attention results pass through the feed-forward network (similar to a
conventional
Transformer network) to produce the final output of the current complementation layer:
\begin{equation}
    X^{i}_{\modala} = \text{LN}(\tilde{X}^i_{\modala} + \text{FFN}(\tilde{X}^i_{\modala})).
    \label{eq:X_final}
\end{equation}


\begin{figure*}[ht]
\centering
\includegraphics[trim=1.2cm 0.5cm 1.2cm 0.8cm,clip, width=0.8\textwidth]{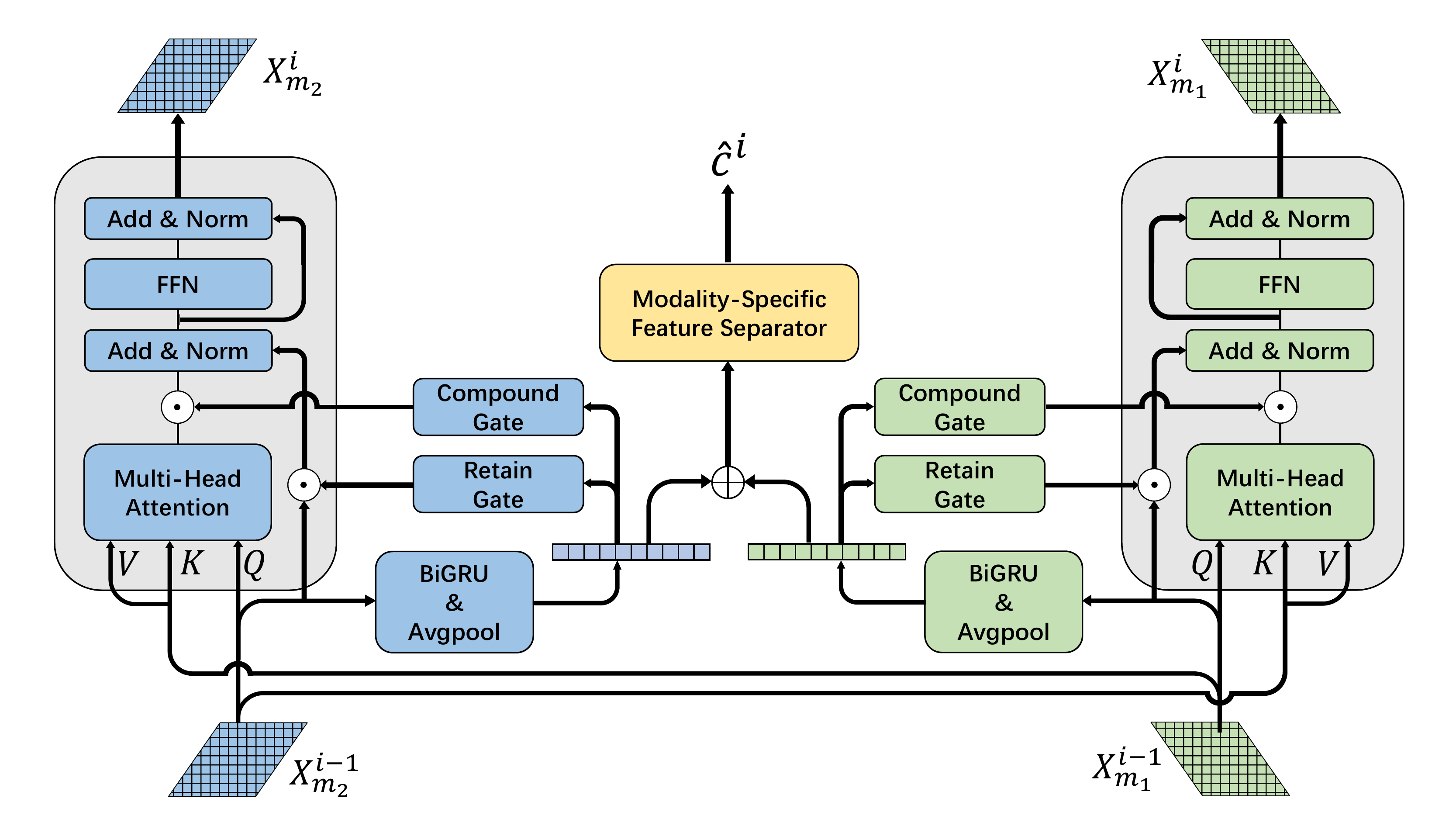}
\caption{A single complementation layer: two identical pipelines (left and right) propagate the \modalafull\ modality and fuse that with \modalbfull\ modality with regularization and gated control.}
\label{Fig.SingleLayer}
\end{figure*}

\subsection{Output Layer and Training}

According to~\eqref{eq:X_final}, a given complementation layer produces the output $X^i_m$, where, similarly to~\eqref{eq:X}, $X^i_m=(x^i_{m,0}, \dots,x^i_{m,n})$. 
When speaking of a layer of a specific complementation module $M$, where $M\in\{TA,TV,VA\}$, we will add $M$ as index: $X^i_{M,m}=(x^i_{M,m,0}, \dots,x^i_{M,m,n+1})$.
The final output of the module $M$ for the modality $m$ is $X^L_{M,m}$, where $L$ is the number of layers; see Fig. \ref{Fig.GTR}.

We compute the final representation by first extracting the heads $\mathbf{h}^{L}_{M,m}=x_{M,m,0}^{L}$ from the outputs of the last layer in each module and then concatenating them.
We have also tried other methods such as average pooling and LSTM or GRU and found them producing similar results, so we chose the most computationally efficient one. As there are two heads in each of the two multimodal complementation layers, concatenating all the outputs of these
four heads in total gives the final representation $\mathbf{h}_{\text{final}}\in \mathbb{R}^{4d}$, where the dimension of each head output is $d$.
Finally, the representation vector is fed to a feed-forward network to produce the final prediction $\hat{y}$.


The loss function comprises two parts: the task loss and the sum of all separators' classification loss. 
In the case of sentiment intensity prediction, the task loss is the mean squared error (MSE), since it is a regression problem. 
In the case of humor detection, we used binary cross-entropy (BCE) loss to facilitate the training for this binary classification task. 
Separator loss is a layer-wise loss so we sum up the results that are computed in each layer and add them to the total loss. 
The total loss is calculated as
\begin{equation*}
\begin{split}
    \mathcal{L} &= \frac{1}{N_b}\sum_{j=1}^{N_b}\left(\tau(y_j,\hat{y}_j)+ \frac{\lambda K}{2L} \sum_{i=1}^L\sum_{M} \mathcal{L}_{sep}^{M,i}\right),
\end{split}
\end{equation*}
where 
$\tau$ denotes the task loss and $\lambda$ is a tunable parameter to control the power of regularization.

\section{Experiments}
\subsection{Datasets}
We evaluated our \modelname\ model on two tasks: sentiment intensity prediction and humor detection, with three datasets involved. 


\begin{itemize}
\item 
\textit{\textbf{CMU-MOSI}}. The CMU-MOSI dataset~\citep{zadeh2016multimodal} is a prevalent benchmark for evaluating fusion networks' performance on the sentiment intensity prediction task. 
The dataset is composed of many YouTube video blogs or vlogs, in which a speaker narrates their opinions on some topic. 
It contains 2,199 utterance-video segments sliced from 93 videos played by 89 distinct narrators. Each segment is manually annotated with a real number score ranged from $-3$ to $+3$, indicating the relative strength of negative (score below zero) or positive (score above zero) emotion. 

\item
\textit{\textbf{CMU-MOSEI}}. The CMU-MOSEI dataset~\citep{zadeh2018multimodal}~is an upgraded version of CMU-MOSI concerning the number of samples. It is also enriched in terms of the versatility of speakers and covers a broader scope of topics. The dataset contains 23,453 video segments, which are annotated in the same way as CMU-MOSI. These segments are extracted from 5,000 videos involving 1,000 distinct speakers and 250 different topics.

\item
\textit{\textbf{UR-FUNNY}}. UR-FUNNY~\citep{hasan2019ur}~is a popular humor detection dataset, as our test benchmark.
This dataset contains 16,514 multimodal punchlines sampled from the TED talks. Each sample 
is annotated with an equal number of binary labels indicating if the protagonist in a video expresses a sort of humor. 
\end{itemize}

\subsection{Preprocessing}
To produce machine-understandable inputs for our model and ensure fair competition with other baselines, following many previous works we process data from the three modalities into typical tensors as introduced below.

\paragraph{\Tm\ Modality}
Many previous works adopted~\citep{pennington2014glove} as word embedding sources. But recent works including current SOTA preferred advanced pretrained language models. Therefore as we stated before in Section \ref{BERT} to encode input raw text in all experiments.

\paragraph{\Vm\ Modality}
Specifically, for experiments on CMU-MOSI and CMU-MOSEI, we use Facet, an analytical tool built on the Facial Action Coding Systems (FACS)~\citep{ekman1997face} to extracted facial features. For experiments on UR-FUNNY we use another facial behavioral analysis tool, Openface~\citep{baltruvsaitis2016openface} to capture facial gesture variance of every speaker.
The resulting vector lengths for the three datasets (MOSI, MOSEI and UR-FUNNY) are 47, 35 and 75 respectively.

\paragraph{\Am\ Modality}
\Am\ features were extracted with COVAREP~\citep{degottex2014covarep}, a professional \am\ analysis framework.

\paragraph{Modality Alignment}
The input signals in our experiments were word-aligned. 
Following many previous works~\citep{tsai2019multimodal, pham2019found, zadeh2018multi}, we used P2FA~\citep{yuan2008speaker} to align \vm\ and \am\ signals to the same resolution of \tm. The tool automatically separates numerous frames into several groups and match each group with a token by averaging their representation vectors to a new one. We used BERT-base-uncased as the text embedding source for all models in all experiments.

\subsection{Baselines and Metrics}
We compared our results with several advanced multimodal fusion frameworks:
\begin{itemize}
    \item
\textbf{DFF-ATMF}~\citep{chen2019complementary}: It is the first bimodal model which first learns individual modality features then executes attention-based modality fusion.
    \item
\textbf{Low-rank Matrix Fusion (LMF)}~\citep{liu2018efficient}: It decomposes high-order tensors into many low-rank factors then performs efficient fusion based on these factors.
    \item
\textbf{Tensor Fusion Network (TFN)}~\citep{zadeh2017tensor}: This approach models intra-modality and inter-modality dynamics concurrently with local feature extraction network and 3-fold Cartesian product. 
    \item
\textbf{Multimodal Factorization Model (MFM)}~\citep{tsai2019learning}: To enhance the robustness of the model of capturing intra- and inter-modality dynamics, MFM is a cycle style generative-discriminative model.
\item
\textbf{Interaction Canonical Correlation Network (ICCN)}~\citep{sun2020learning}: Correlation between modalities is a latent trend to be excavated under the fusion process. ICCN purely relies on mathematical measure to accomplish the fusion process.
\item
\textbf{MulT}~\citep{tsai2019multimodal}: To alleviate the drawback of hard temporal alignment for multimodal signals, MulT utilizes stacked transformer networks to perform soft alignment to extend the number of positions on the time axis that each frame of signal can interact with.
\item
\textbf{MISA}~\citep{hazarika2020misa}: Motivated by previous work in domain separation task, this work regards signals from different modalities as data in different domains and similarly constructs two kinds of feature spaces to finish the fusion process.
\end{itemize}

We used five different metrics to evaluate the performance on CMU-MOSI and CMU-MOSEI: mean absolute error (MAE),  which directly calculates the error between predictions and real-number labels; seven-class accuracy (Acc-7), positive/negative (excluding zero labels) accuracy (Acc-2) and F1 score, which coarsely estimate the model's performance by comparing with quantified values; and Pearson correlation (Corr) with human-annotated truth, which measures standard deviation. 
As for humor recognition on UR-FUNNY, it is a binary classification problem and we only report the binary classification accuracy (Acc-2).

\section{Results and Analysis}
\subsection{Summary of Results}
We list the results with baselines on all the three datasets in Table~\ref{mosi,mosei}.
From these results, it can be found that our \modelname\ outperforms other models on almost all metrics (except for the correlation coefficient on CMU-MOSI). 
We attained an improvement of around 1\% over the state-of-the-art approaches in terms of binary classification accuracy and more than 2.5\% in terms of 7-class accuracy.
We attribute the difference partly to the insensitivity of coarse metrics to the variation in the model's predictions. 
The best performance boost, more than 4\%, was on the MAE of CMU-MOSEI.

\begin{figure}[th]
    \centering
    \includegraphics[width=0.45\textwidth, trim=0.1cm 15cm 0.1cm 0cm]{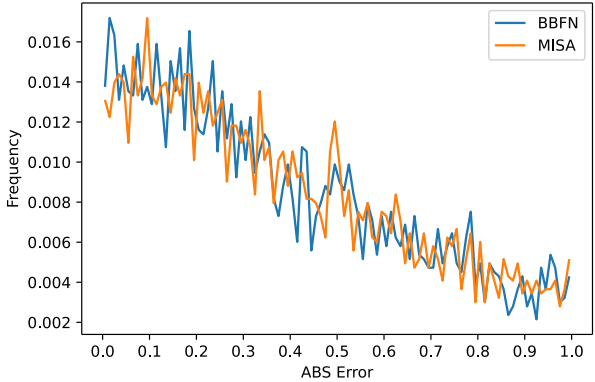}
    \caption{Distribution of absolute error when testing \modelname~and MISA on CMU-MOSEI dataset.}
    \label{Error Distribution}
\end{figure}

To better illustrate how \modelname\ beats the SOTA (MISA), we compute the absolute errors of all the predictions on the test set of CMU-MOSEI and paint their distributions in Fig. \ref{Error Distribution}. It can be observed that in the low error part (error$<0.25$) the curve of \modelname\ has more peaks than MISA, which demonstrates the higher precision that our model can reach.

\begin{table*}[ht!]
    \centering
    \resizebox{0.8\linewidth}{!}{
    \begin{tabular}{l|c c c c c|c c c c c|c}
        \toprule
        \multirow{2}{6em}{Models} & \multicolumn{5}{c|}{CMU-MOSI} & \multicolumn{5}{c|}{CMU-MOSEI\rule[-1.5ex]{0ex}{0ex}} & UR-FUNNY\\
        ~ & MAE & Corr & Acc-7 & Acc-2 & F1 & MAE & Corr & Acc-7 & Acc-2 & F1 & Acc-2 \\
        \midrule
        $\text{DFF-ATMF}^{\triangle}$ & - & - & - & 80.9 & 81.2 & - & - & - & 77.1 & 78.3 & - \\
        $\text{LMF}^{\triangle}$ & 0.917 & 0.695 & 33.2 & 82.5 & 82.4 & 0.623 & 0.677 & 48.0 & 82.0 & 82.1 & 67.53 \\
        $\text{TFN}^{\triangle}$ & 0.901 & 0.698 & 34.9 & 80.8 & 80.7 & 0.593 & 0.700 & 50.2 & 82.5 & 82.1 & 68.57\\
        $\text{MFM}^{\triangle}$ & 0.877 & 0.706 & 35.4 & 81.7 & 81.6 & 0.568 & 0.717 & 51.3 & 84.4 & 84.3 & - \\
        $\text{ICCN}^{\triangle}$ & 0.862 & 0.714 & 39.0 & 83.0 & 83.0  & 0.565 & 0.713 & 51.6 & 84.2 & 84.2 & - \\
        $\text{MulT}^{\dag}$ & 0.832 & 0.745 & 40.1 & 83.3 & 82.9 & 0.570 & 0.758 & 51.1 & 84.5 & 84.5 & 70.55 \\
        $\text{MISA}^{\dag}$ & 0.817 & 0.748 & 41.4 & 82.1 & 82 & 0.557 & 0.748 & 51.7 & 84.9 & 84.8 & 70.61 \\
        \midrule
        $\text{\modelname}^{\ddag}$  (Ours) &\textbf{0.776} & \textbf{0.755} & \textbf{45.0} & \textbf{84.3} & \textbf{84.3} & \textbf{0.529} & \textbf{0.767} & \textbf{54.8} & \textbf{86.2} & \textbf{86.1} & \textbf{71.68} \\
        \bottomrule
    \end{tabular}
    }
    \caption{Results on the test set of CMU-MOSI and CMU-MOSEI dataset. Notation: $\triangle$ indicates results in the corresponding line are excerpted from previous papers; $\dag$ means the results are reproduced with publicly visible source code and applicable hyperparameter setting; $\ddag$ shows the results have experienced paired t-test with $p<0.05$ and demonstrate significant improvement over MISA, the state-of-the-art model.} 
    \label{mosi,mosei}
\end{table*}

\subsection{Ablation Study}
To examine the functionality of the overall architecture and 
the
components introduced in this work, we conducted an ablation study on CMU-MOSEI dataset; see \ref{ablation}.

\begin{table*}[ht!]
  \small
    \centering
    \begin{tabular}{c lll | c c c}
        \toprule
        \multirow{2}{*}{Case} & \multicolumn{3}{c|}{Input by Modality} & \multicolumn{3}{c}{Prediction \& Truth\rule[-1.5ex]{0ex}{0ex}} \\
        & \multicolumn1c\Tm & \multicolumn1c\Vm & \multicolumn1{c|}\Am & Prediction & Truth & ABS Error \\
        \midrule
       \multirow{2}{*}{A} & \multirow{2}*{\shortstack[l]{\it But, I mean, if you're going to watch a movie like that, go see Saw \\ \it again or something, because this movie is really not good at all.}} & \multirow{2}{*}{\shortstack[l]{ Widely opened \\ eyes }} & \multirow{2}{*}{\shortstack[l]{Pause and Stress}}  & \multirow{2}{*}{$-1.973$} & \multirow{2}{*}{$-2.000$} & \multirow{2}{*}{$0.027$} \\
       & & & & & & \\
       \midrule
       \multirow{2}{*}{B} & \multirow{2}*{\shortstack[l]{\it Plot to it than that the action scenes were my favorite parts \\ \it  though it's.}} & \multirow{2}{*}{\shortstack[l]{Smiling face \\ Relaxed wink}} & \multirow{2}{*}{\shortstack[l]{Stress \\ Pitch variation}} & \multirow{2}{*}{$+1.638$} & \multirow{2}{*}{$+1.666$} & \multirow{2}{*}{$0.028$} \\
      & & & & & & \\
       \midrule
       \multirow{2}{*}{C} & \multirow{2}*{\shortstack[l]{\it (umm) So if you're looking for something it's sort of lighthearted.  }} & \multirow{2}{*}{\shortstack[l]{No expression}} & \multirow{2}{*}{\shortstack[l]{Normal Voice \\ Peaceful tone}} & \multirow{2}{*}{$-0.016$} & \multirow{2}{*}{$0.000$} & \multirow{2}{*}{$0.016$} \\
      & & & & & & \\
        \bottomrule
    \end{tabular}
    \caption{Input and predictions of two samples in our case study.}
    \label{Case Study}
\end{table*} 

In five experiments we verified the effect of the bimodal fusion architecture.
Specifically, we 
(1)~only used one pair as input; 
(2)~replaced input \tm-related modality pairs (TV, TA) with \vm/\am-related ones;
and 
(3)~added a \vm-\am\ complementation module to make a ternary symmetric model. In all cases, separators and gates were used.

The models of type~(2) 
outperformed those of type~(1) on MAE and Acc-7 (the most accurate measures), which indicates that all three modalities are important.
Moreover, 
the performance of \vm-focused input (TV+VA) is 
close to that of \tm-focused input (our TV+TA), i.e., 
our architecture can operate on these modality pairs, too.

On the other hand, the performance degrades on type~(3), when \vm-\am\ input pairs are added. 
That is, even after including all modalities in the input, redundant network architecture can cause harmful effects bringing in malicious noise, which damages collected useful information and confuses the model. 

We also explored the benefits of the feature-space separator and gated control by 
removing the separator, the two gates, or both from our \modelname. The outcome shows some degradation in all metrics except the correlation. 
This proves that including gated control and modality separator improves the model's performance, 
though the greatest improvement over the baselines shown in Table \ref{ablation} comes from our overall bi-bimodal architecture.

\begin{figure}[th]
    \centering
    \includegraphics[trim=0.1cm 0cm 0.1cm 0cm, width=\columnwidth]{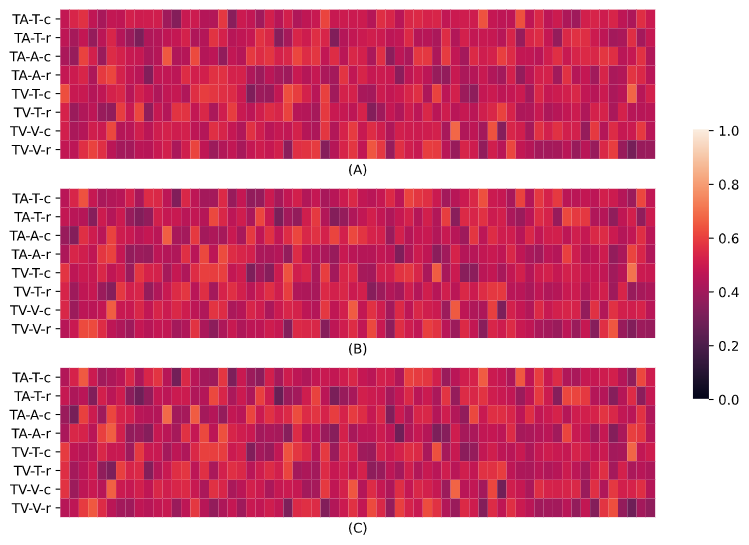}
    \caption{Visualization of eight gated control signals in the second layer of our \modelname\ for two case study samples. ``XY-X/Y-c/r'' denotes the compound/retain gate in the transformer pipeline for X/Y modality in XY complementation module.}
    \label{Gate Weights}
\end{figure}

\subsection{Further Analysis}
To study how the gates affect the information flow, we visualized the weights in all the gates per dimension; see Fig. \ref{Gate Weights}.
We hypothesize that the discrepancy in weight distributions reflects the relative importance of modalities.
Specifically, for the two gates associated with one modality in the same module, if the weights in the compound gate are greater than those in the retain gate, it implies that the model enforces the corresponding modality to learn much from its counterpart in the module and the modality thus is less important. Conversely, if the weights of the retain gate are greater, then the modality is more important than its counterpart.
Fig. \ref{Case Study} shows three typical samples, including raw data input (for \vm~and~\am~we only give short descriptions), predictions and truths from the test set.
In case A, most cues are attained from the \tm\ modality to express sort of disappointment, while data from the \vm\ and \am\ modalities are not so informative. 
Hence for the \am\ and \vm, the model makes the corresponding modalities A and V pay more attention to the heterogeneous attention results instead of themselves, which is indicated by the larger weights in the compound gates of the two modalities (TA-A-c \& TV-V-c).

In case B, the V and A modalities are seen to be providing key information along with T. Therefore, the weight difference is indiscernible, and two paths of information flow achieve a balance. 
Surprisingly, Fig.~\ref{Gate Weights} shows that for the text modality in the TV complementation module, the weights in the compound gate are slightly higher than those in the retain gate. 
This implies that the textual modality can be complemented by the information attained from the visual modality.

In case C, no single modality can provide clear evidence for the neutral sentiment, but each of them serves as a favorable supplementary to others. Therefore, in Fig.~\ref{Gate Weights} we find the weights in the gates of each modality are comparable, indicating the similar dependency of bimodal fusion results on both modalities.

We also compared our \modelname\ with MISA for the final prediction in the two cases. As shown in Table \ref{Case Study}, \modelname's outputs are closer to the ground truth, owing to the fine-grained control offered by these gates, whereas MISA makes opposite (case~A) or conservative (case~B) predictions because, as a result of ternary-symmetric architectures, it is distracted by insignificant modalities, which add disturbing factors and dilute the pertinent features. 

\begin{table}[t]
    \centering
    \resizebox{\ifdim\width>\columnwidth\columnwidth\else\width\fi}{!}{
    \begin{tabular}{l|c c c c c}
        \toprule
        Description & MAE & Corr & Acc-7 & Acc-2 & F1 \\
        \midrule
        TV only & 0.546 & 0.761 & 51.8 & 85.6 & 85.6 \\
        TA only & 0.548 & 0.759 & 51.7 & 85.5 & 85.5 \\
        VA only &  0.816 & 0.261 & 41.1 & 71.1 & 64.5 \\ \midrule
        VA+TA & 0.533 & 0.773 & 54.1 & 84.8 & 84.9  \\ 
        TV+VA & 0.531 & \textbf{0.775} & 54.5 & 85.7 & 85.7 \\
        TV+TA (\modelname) & \textbf{0.529} & 0.767 & \textbf{54.8} & \textbf{86.2} & \textbf{86.1} \\
      \quad  w/o separator & 0.533 & 0.766 & 54.1 & 85.7 & 85.4 \\
      \quad  w/o gates & 0.531 & 0.768 & 53.9 & 85.8 & 85.7 \\
      \quad  w/o both  & 0.540 & 0.763 & 53.0 & 85.1 & 85.0 \\ \midrule
        TV+TA+VA & 0.547 & 0.768 & 52.8 & 84.3 & 84.4 \\ 
        \bottomrule
    \end{tabular}
    }
    \caption{An ablation study of \modelname's architecture and functional components on the test set of CMU-MOSEI.}
    \label{ablation}
    \vspace{-3em}
\end{table} 

\section{Conclusion}
We have presented \modelfull\ (\modelname), a fusion architecture for  multimodal sentiment analysis 
that
focuses on bimodal fusion process.
Pairwise fusion process proceeds progressively through stacked complementation layers in each learning module. 
To alleviate the issue of feature space collapse and lack of control at fusion time, we introduced into our model the structure of modality-specific feature space separator and gated control mechanism, respectively.
Comprehensive experiments and analysis show that our model outperforms the current state-of-the-art approaches. Despite 
good
performance of our model, we plan to explore 
more advanced fusion methods and architectures.
Also besides sentiment analysis, in multimodal research there are many other important tasks, for which we can combine task-specific techniques with appropriate fusion schemes. 
Accordingly, we plan to improve the fusion quality of multimodal data as well as the coordination of fusion and task solving modules.

\section*{Acknowledgements}
This material is based upon work partially supported by the SUTD SRG grant \#T1SRIS19149, National Science Foundation (Awards \#1722822 and \#1750439), and National Institutes of Health (Awards \#R01MH125740, \#R01MH096951, \#U01MH116925, \#U01MH116923). Any opinions, findings, conclusions, or recommendations expressed in this material are those of the author(s) and do not necessarily reflect the views of the National Science Foundation or National Institutes of Health, and no official endorsement should be inferred.

\bibliographystyle{ACM-Reference-Format}
\bibliography{ACMref.bib}

\end{document}